\documentclass{article}
\usepackage[utf8]{inputenc}
\usepackage[margin=1in,footskip=0.5in]{geometry}
\usepackage{times}  
\usepackage{helvet} 
\usepackage{courier}  
\usepackage[hyphens]{url}  
\usepackage{graphicx} 
\usepackage[switch]{lineno}
\usepackage{hyperref}
\usepackage{bm,amsmath,amsthm,amssymb,amsfonts}
\long\gdef\affiliations #1{ \def \affiliations_{#1}}
\newcommand{\norm}[1]{\left\lVert#1\right\rVert}

\usepackage{hyperref}
\usepackage{url}
\usepackage{graphicx}
\usepackage{bm,amsmath,amsthm,amssymb,amsfonts}
\usepackage{xcolor}

\title{Removing Racial and Gender Bias in TED Talk Ratings by Awareness of Verbal and Gesture Quality}

\author {

        Ankani Chattoraj \thanks{equal contribution},\textsuperscript{\rm 1} 
        Rupam Acharyya \footnotemark[1], \textsuperscript{\rm 1}
        Shouman Das \textsuperscript{\rm 1}
         Md. Iftekhar Tanveer  \textsuperscript{\rm 2}
          Ehsan Hoque \textsuperscript{\rm 1} \\
}

\date{
     University of Rochester\textsuperscript{\rm 1}, 
    Spotify Research \textsuperscript{\rm 2} \\
    achattor@ur.rochester.edu, 
    racharyy@ur.rochester.edu, sdas13@ur.rochester.edu, go2chayan@gmail.com, mehoque@gmail.com
}

\begin{document}
\maketitle
\begin{abstract}
The role of verbal and non-verbal cues towards great public speaking has been a topic of exploration for years. We identify a commonality across present theories, the element of ``variety or heterogeneity'' in modes of communication (e.g. resorting to stories, scientific facts, emotional connections, facial expressions etc.) which is essential for effectively communicating information. Based on this observation, we formalize a novel HEterogeneity Metric, HEM, that quantifies the quality of a talk both in the verbal and non-verbal domain (transcript and facial gestures). Using TED talks as an input repository of public speeches, we show that there is a meaningful relationship between HEM and the ratings of TED talks given to speakers by viewers. Further, we discover that HEM successfully captures the prevalent bias in ratings w.r.t race and gender. We incorporate the HEM metric into the loss function of a neural network and show improvement of fairness in rating prediction. Our work ties together a novel metric for public speeches in both verbal and non-verbal domain to design a fair rating prediction system.
\end{abstract}

\section{Introduction}

\begin{figure}
    \centering
    \includegraphics[scale=0.8]{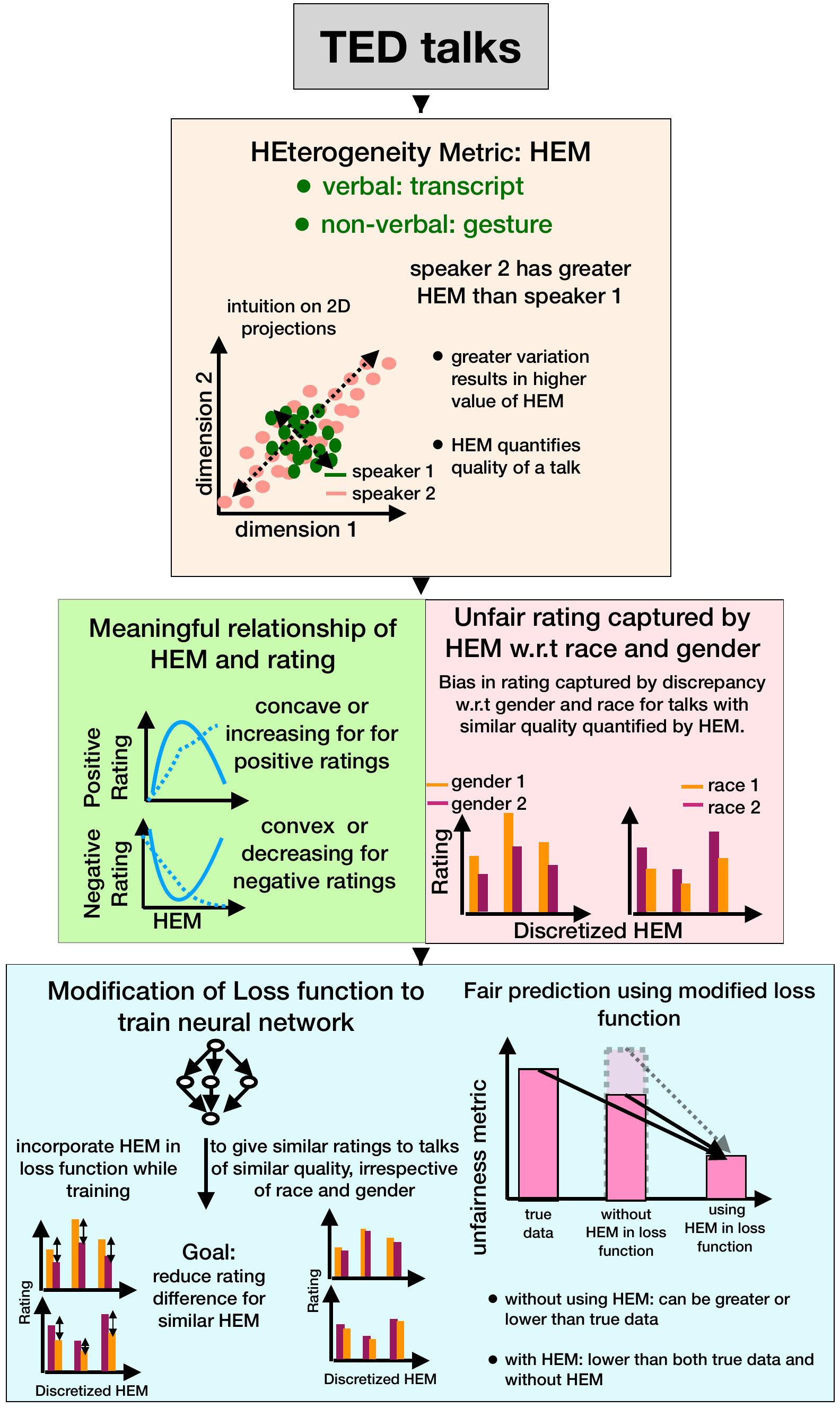}
    \caption{\textbf{Fair prediction pipeline using $HEM$ metric.} Step 1: Define heterogeneity based $HEM$ for both verbal and non verbal aspects of a talk. Step 2: Show meaningful relationship between $HEM$ and rating labels. Step 3: Show $HEM$ captures bias in rating w.r.t gender and race. Step 4: Train a neural network with $HEM$ in loss function to obtain fair rating predictions} 
    \label{fig:pipeline}
\end{figure}
A good speaker efficiently expresses inner thoughts to an audience \cite{costello1930psychological, newcomb1917educational}, often encompassing ethos, pathos and logos \cite{sep-aristotle-rhetoric}, adding humor, asking questions, using quotations, drawing analogies etc \cite{sandmann2013introductions,fisher1989human, garlick1993verbal, bitzer1999rhetorical,ting2018communicating,decaro2012audience}. For example, celebrity chef Jamie Oliver gave an award winning TED speech in 2010
where he attempted to convince a diverse audience to change their most basic eating habits. After establishing credibility, he showed compelling statistics confirming that death from diet related disease is more prevalent than any other diseases, accidents or murder. He ended with easy alternative solutions to encourage healthy eating \cite{oliver2010teach}. Different aspects of the talk resonated with different people, depending on their age, life style, experiences and cultural background, making it an awarding winning speech. In addition to verbal, nonverbal components of a talk also play a key role in determining its appeal \cite{lucas2004art}. Effective  use  of both transcript and gesture and a deliberate alteration of message through verbal and nonverbal cues give shape to the main message of a speech \cite{abbott2008cambridge,tanveer2018awe}. \\
One common thread that ties the important aspects of a good public speech in both verbal and non-verbal domain is ``variety or heterogeneity'': the efficient way of conveying the main message by providing information in the form of stories, scientific facts, emotional connections, personal experience etc. We formalize this by defining a novel, ``HEtero-geneity Metric'' ($HEM$). We conduct our investigation on a diverse set of talks and ratings as found in the TED talk website. The speakers and the viewers are from different cultures, age groups, and backgrounds. We show that $HEM$ has a meaningful relationship with the ratings of TED talks in both the verbal and non verbal domain. We find that the desirable (positive rating, e.g., fascinating) and undesirable (negative rating, e.g., unconvincing) ratings of talks grow and shrink respectively with the increase in $HEM$, for both the verbal and non verbal domain.\\
Interestingly, we also find discrepancies in rating of TED talks w.r.t race and gender for similar $HEM$. Motivated by this observation, we introduce \emph{fairness by quality} in designing a public speech rating prediction model. Our prediction model is a neural network with a modified loss function which aims to reduce discrepancy in ratings for talks with comparable quality as quantified by $HEM$ besides improving prediction accuracy. 

\section{Related work}
Public speaking has been used as a tool to reach out to masses for centuries. A good talk requires credibility or trustworthiness, emotional appeal, logic or reasoning etc \cite{barrett1987sophists, sep-aristotle-rhetoric}. In recent years, with a more multi-cultural and age diverse audience \cite{dhanesh2012speaking, ting2018communicating,decaro2012audience}, it has been proposed that adding humor, asking a question, using analogies and quotations can be effective ways of connecting to the audience \cite{neville2002exploration,xiao1996hierarchical,lustig2006intercultural,carlson2005wow,scholl2013special}. 
Research also shows that majority of all human communication is non-verbal \cite{neuliep2020intercultural,mehrabi2019survey,mehrabian1981silent} and basic facial expressions of emotion are similar across cultures \cite{ekman1997face}. For example we typically smile while speaking about happy events, we open our eyes wide with astonishment, we raise an eyebrow in suspicion or open our mouth to express surprise, we pucker our lips to express disapproval, disappointment or concern. Because of such connections between emotions and facial expression, they form an integral part of delivering a good speech \cite{monroe1974principles, giffin1971fundamentals,gregory2010public,sprague1984tlie,ball1984sense}.\\
We stitch together all these aspects of a good speech by their ``variety and heterogeneity'' both in the verbal and non-verbal domain. However, too much heterogeneity in transcript can be detrimental because it causes cognitive overload (``bounded rationality'') \cite{simon1982models}. Similarly, too much variability in facial expressions can be interpreted as nervous gestures (Chapter 12 of \cite{schreiber2017introduction}). \\
Mitigation of bias in AI models has drawn much interest in recent times as automated systems are now being used in sensitive decision making such as criminal justice systems \cite{tollenaar2013method}, filtering resumes for job applicants \cite{nguyen2016hirability, chen2017automated, naim2016automated} etc. 
A recent survey about fairness in machine learning is very well captured in \cite{mehrabi2019survey}. However, there has not been much work of incorporating fairness in the domain of public speech rating prediction. In a recent work \cite{acharyya2020fairyted} the authors used counterfactual fairness \cite{kusner2017counterfactual} for predicting the rating of public speaking in TED talks. Their work lacks explainability and focuses on fair predictions in the verbal domain only. We take a complementary approach of first introducing a psychology motivated intuitive metric ($HEM$) for both verbal and non verbal domain of a talk and then using it to identify the existence of bias in TED talk ratings and finally mitigating it. 

\section{Data}\label{data-section}
\subsection{Data Acquisition:} We collected the public speaking data from TEDtalk website~\footnote{Videos on \url{www.ted.com}}. The dataset contains speeches published between 2006 and 2017 from $1980$ talks (after removal of outliers) covering $>400$ categories such as science, technology, global issues, health, business entertainment etc. These talks have millions of viewers \footnote{More than 12 million subscribers on YouTube \url{https://www.youtube.com/user/TEDtalksDirector}} around the world who come from various background, age and culture. These viewers rate the talks after viewing them based on multiple labels of which we consider $3$ positive/desirable rating labels: \textit{fascinating, ingenious} and \textit{jaw-dropping} and $3$ negative/undesirable rating labels: \textit{long-winded, unconvincing} and \textit{ok}. We also collected data on the sensitive attributes $S$ (\textit{race} and \textit{gender}) using Amazon mechanical turk following \cite{acharyya2020fairyted}. This was done to mimic how viewers identify the sensitive attributes and no authors were involved in assignment of the values of sensitive attributes. We use information about the data attributes $X$: \emph{transcript ($Tr$), visual gestures ($V$), view counts ($C$), sensitive attributes $S$: race ($R$), gender ($G$) and rating label ($Y$)} of each talk to design our prediction model. Summary of the dataset is given in Table \ref{tab:datasize} in Appendix. 
\subsection{Extracted Features} 
\textbf{Verbal:} We obtain $HEM_{tr}$ by obtaining doc2vec~\cite{le2014distributed} representation of each transcript ($Tr \in \mathbb{R}^d$) using \texttt{gensim} library~\cite{rehurek_lrec} and set $d = 200$.\\
\textbf{Non-verbal:} We obtain $HEM_{ges}$ by using OpenFace \cite{baltruvsaitis2016openface} to extract 17 facial action units. These features (every 30 frames/sec), $V \in \mathbb{R}^d$, where $d = 17$, are based on the Facial Action Coding System (FACS) \cite{ekman1997face}.
\subsection{Data Preprocessing} 
We normalize original view counts of the TED talks ($C \in \mathbb{Z}$) using min-max technique to obtain $C \in \mathbb{R}$, making them comparable across attributes.\\
Normalized rating labels $Y$ for each talk was obtained by dividing with its total number of rating counts. This neutralizes the effect a video's online duration, since that is captured by the total views.\\
Binary rating labels were computed by thresholding w.r.t its median value across all talks, $Y^{bin} \in \{0, 1\}$. The attributes of our dataset are in Table \ref{tab:final_data} in Appendix.

\section{HEterogeneity Metric (HEM)} \label{section-hem}
Various theories claim  that a good talk should have credibility, emotional connection to the audience, logical argument, make use of stories, scientific facts, quotations, humor and so on (see Related work) which we call \textit{characteristics}. Each characteristic when represented by words specific to them would vary from each other. We quantify the verbal quality of a talk by formalizing the heterogeneity across \textit{characteristics} prevalent in the transcript ($HEM_{tr}$). 
\begin{figure}[t]
    \centering
    \includegraphics[scale=0.6]{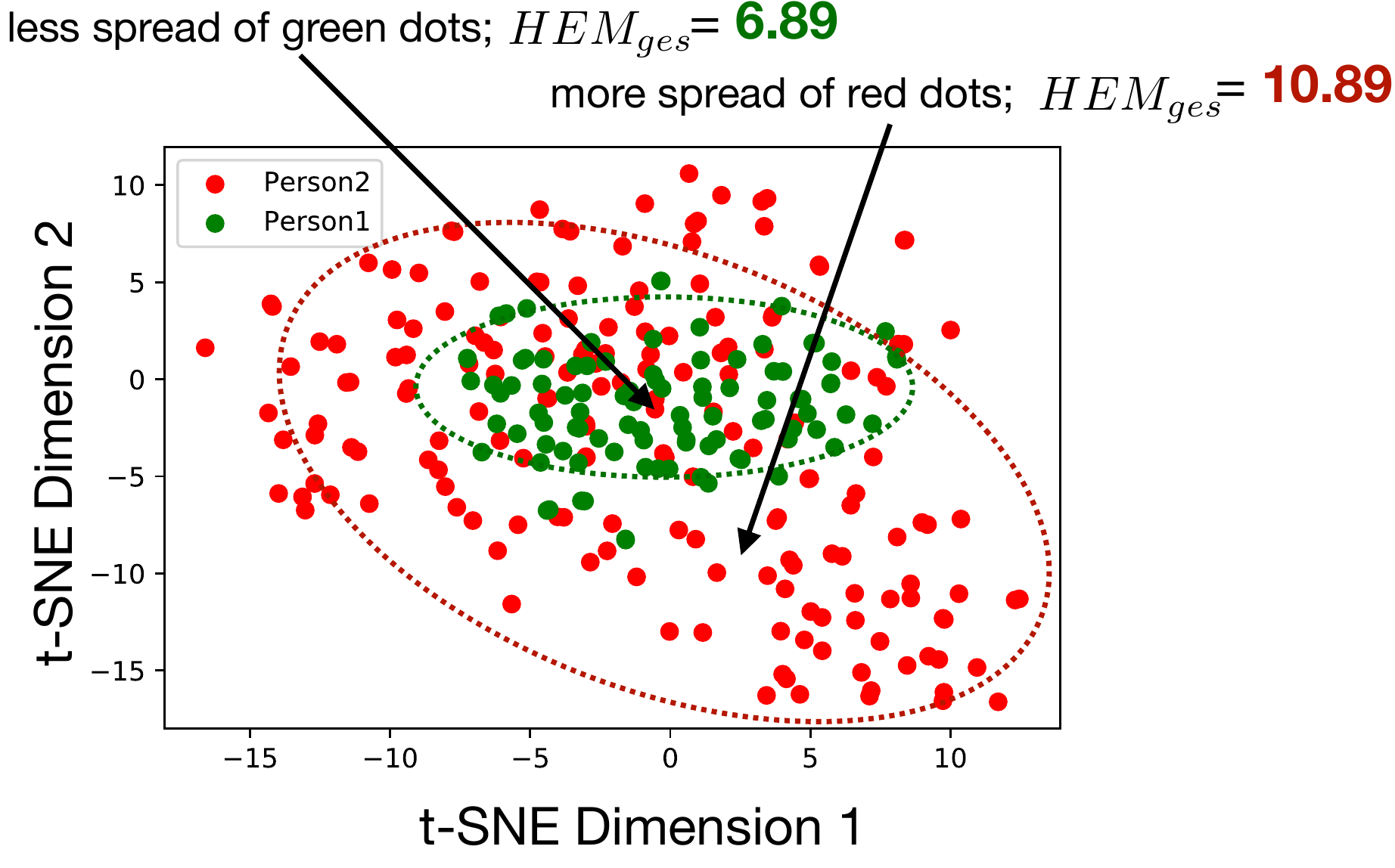}
    \caption{\textbf{Gesture pattern of two random speakers}\\
    Speaker represented with red dots has greater variation and hence greater values of $HEM_{ges}$.}
    \label{fig:visualvariation}
\end{figure}
Further, multiple studies in the domain of gesture analysis show that the lack of variation in gesture and mannerism makes a talk boring. We quantify the non-verbal quality of a talk by the heterogeneity in facial gestures of a speaker ($HEM_{ges}$). It is important to remember that ``quality" here refers to the ``impact and effectiveness" of a speech due to heterogeneity.
\begin{figure*}[t]
    \centering
    \includegraphics[width=\textwidth]{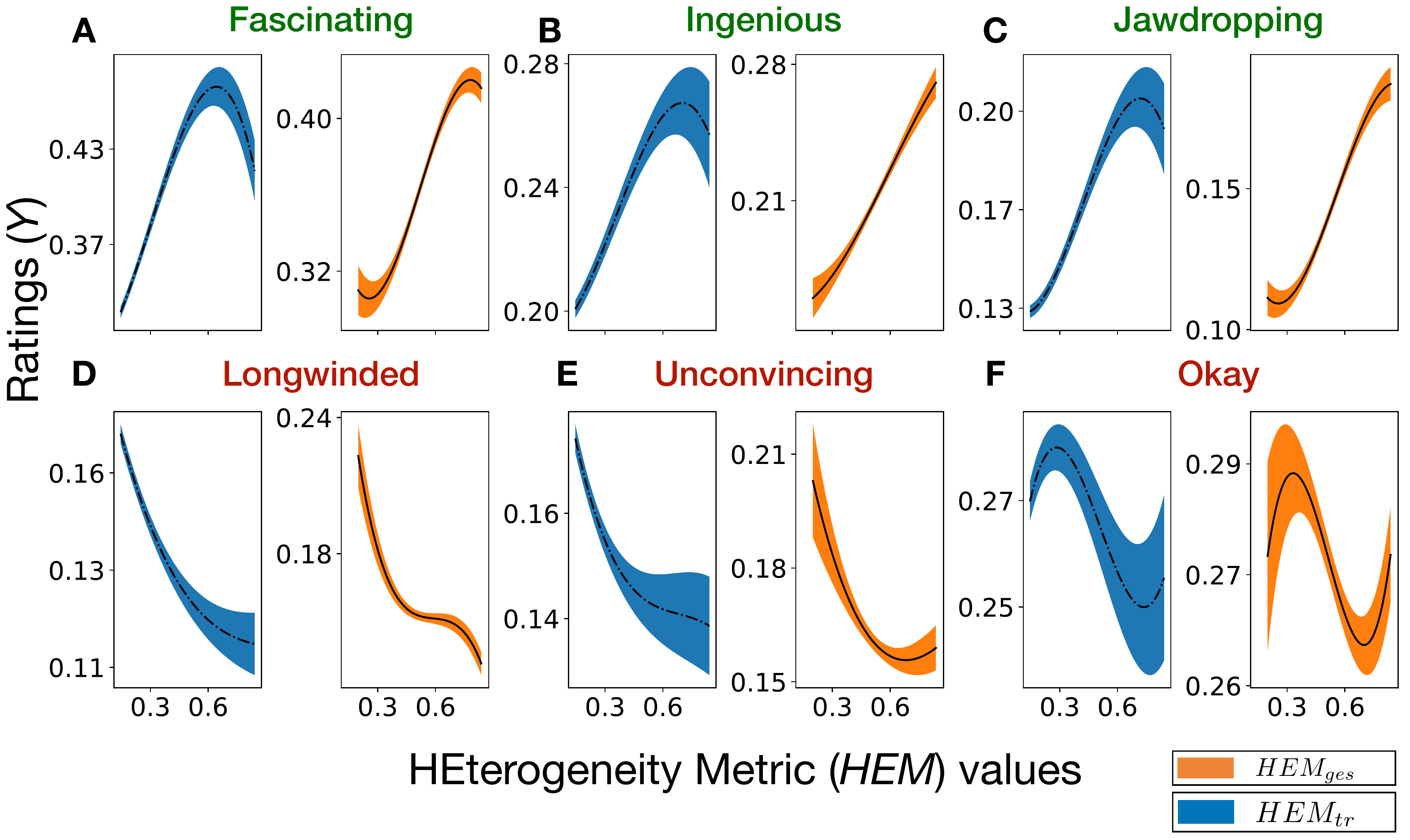}
    \caption{\textbf{$HEM$ metric shows meaningful relationship with rating}\\\textbf{A-C}: Positive ratings show a concave or increasing trend with respect to both $HEM_{tr}$ shown in blue and $HEM_{ges}$ shown in orange. With increase in $HEM$ we expect the quality of speech to improve causing positive rating to increase. However, too much heterogeneity can lead to confusion causing positive ratings to saturate or even decrease. \textbf{D-F}: The trend is opposite for negative rating labels, showing a convex or decreasing relationship.}
    \label{fig:HEM_all}
\end{figure*}
\subsection{Verbal ($HEM_{tr}$)} 
For each talk we obtained $K$ topics using Latent Dirichlet Allocation (LDA)~\cite{blei2003latent}. Each of these topics is a probability distribution over the set of words in the transcript. We define the summary representation of each topic as the weighted combination (weights are the respective probability) of the Glove embeddings \cite{pennington2014glove} for the highest probable words in that topic. We identify these topics as \textit{characteristics} of the talk by using topic modeling within each transcript. The summary representation of $i^{th}$ topic as $\bm{T_i} (\in \mathbb{R}^{300})$ is defined by, 
$\bm{T_i} = \sum_{j=1}^{n}weight(w_j)\cdot glove(w_j),$
where $n$ is the number of highest probable words chosen to represent the topic and $weight(w_j)$ is the weight of the $j^{th}$ word in the $i^{th}$ topic. We then define the representation matrix of all topics as $\bm{T} (\in \mathbb{R}^{K\times 300})$ where the $i^{th}$ row is $\bm{T_i}.$ 
We obtain the topic similarity matrix, $\bm{U}=\bm{T} \bm{T}^\top (\in \mathbb{R}^{K\times K})$ which captures the similarity between topics. 
Heterogeneity metric ($HEM_{tr}$) of a talk is then defined as the product of highest $k$ eigenvalues of $\bm{U}$. This product defines the volume spanned by the most diverse $k$ topics in the topic vector space (see Theorem 5.2 of \cite{kulesza2012determinantal}). Note that, the more diverse the topic vectors, larger will be the magnitude of the product of top $k$ eigenvalues. This would result into larger volume spanned by corresponding topic vectors. Hence more variety and heterogeneity in a talk with respect to the transcript implies greater value of $HEM_{tr}$.  
$HEM_{tr} = \prod_{i=1}^k \lambda_i$ where $\lambda_1\geq \lambda_2\geq \cdots \geq \lambda_K\geq 0$ are the eigenvalues of $\bm{U}$ (all of $\lambda_i$'s are real and non-negative as $U$ is a PSD matrix). In this work, we choose $n = 10, K = 10, k=5$.

\subsection{Non-verbal ($HEM_{ges}$)}
We model the facial gestures using similar technique as in \cite{agarwal2019protecting} which encodes personalized speaking pattern. OpenFace \cite{baltruvsaitis2016openface} toolkit is used to extract 17 facial action units related intensity scores from the video at 30 frames/sec. These intensity scores represent various facial muscle movements like inner brow raiser, outer eyebrow raiser, eyebrow lowerer, upper lid raiser, cheek raiser, jaw drop, eye blink etc. Each talk video is divided into 10 second segments with 5 seconds sliding window. For each of these segments, we calculate Pearson correlation coefficient to quantify the similarity among these intensity scores over time. We calculate a total of $\binom{17}{2}$ = 136 correlation coefficients that encodes facial gesture pattern for each 10 second segments. Facial gesture pattern of two speakers are shown in Figure \ref{fig:visualvariation}. The two colors show gesture pattern from two randomly chosen speakers. Each colored dot is a 2-D t-SNE \cite{maaten2008visualizing} representation of 136 correlation coefficients that encodes facial gesture pattern for a 10 second segment. The spread of dots of each color shows the amount of heterogeneity present in each speaker's facial gesture. Therefore we define the heterogeneity metric for a speaker's gesture pattern $(HEM_{ges})$ as the maximum distance between two points in 136-dimensional space. Intuitively, this maximum distance gives us the measure of highest variability present in facial muscle movements. For example, person 2 denoted by red dots shows a greater spread of 2-D t-SNE values as compared to person 1 denoted by green dots. Correspondingly, the $HEM_{ges}$ for person 2 is 10.89 which is greater than that of  person 1 ($HEM_{ges}$ = 6.89). Let the number of 10 seconds segments in a TED talk video be $s$. Denote $p_{i} (\in \mathbb{R}^{136})$ to be the point representing the correlation coefficients for $i^{th}$ segment. Now, $d_{ij}$ denotes the eucledian distance between points  $p_i$ and $p_j.$ The $HEM_{ges}$ of gesture pattern for the video is defined as, $HEM_{ges} = \max_{i,j \in  \{1,\cdots,s\}} d_{ij}$.

\begin{figure}[t]
    \centering
    \includegraphics[width=\textwidth]{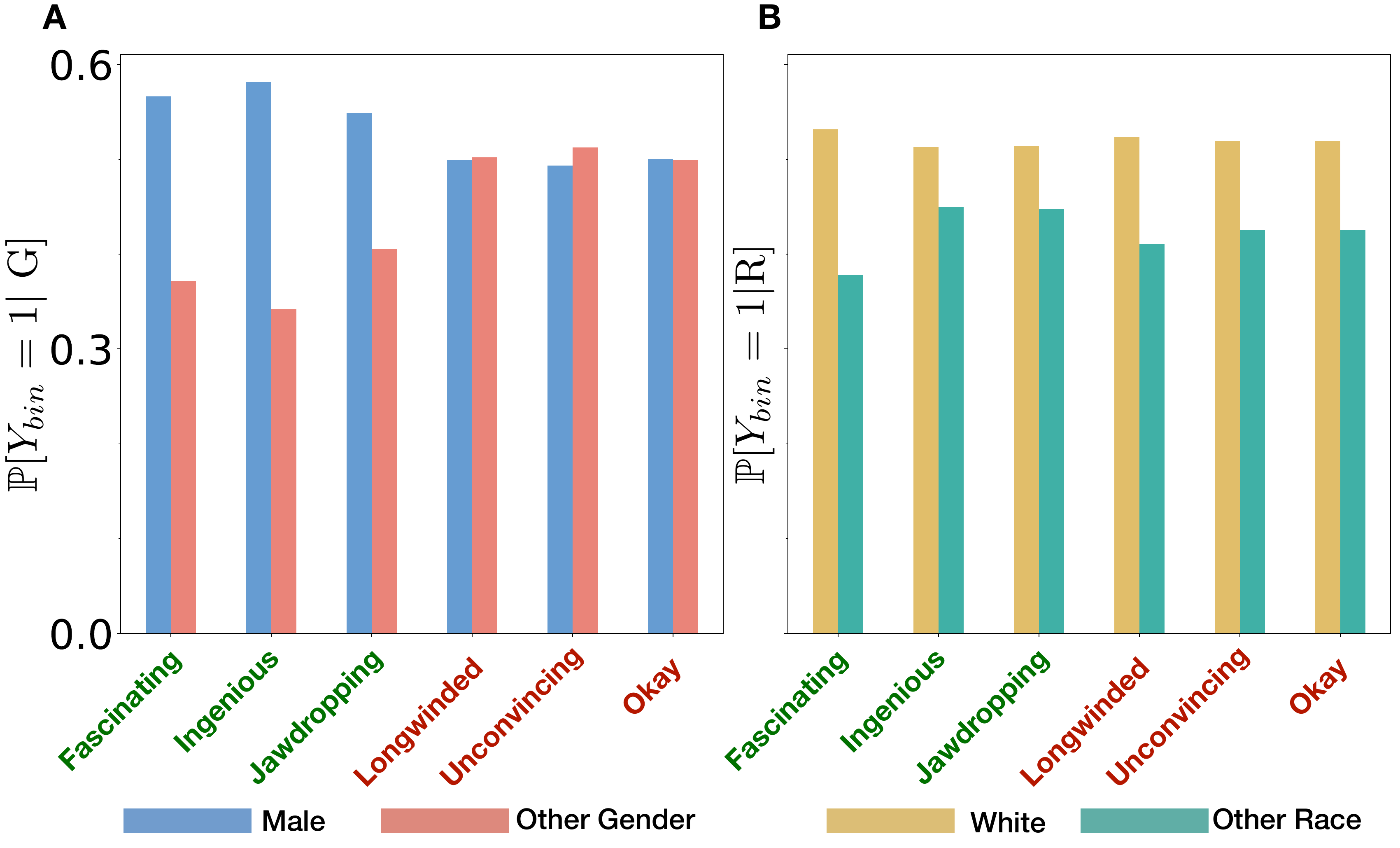}
    \caption{\textbf{Rating bias with respect to gender and race}\\
    \textbf{A:} Male are more likely (significant with $p < 0.05$) to be rated positive labels as compared other gender. No such discrepancy for negative rating labels. \textbf{B:} White speakers are more likely to be rated with all 6 labels as compared to speakers of race. }
    \label{fig:data-bias}
\end{figure}

\subsection{Choice of Rating Labels} 
We assume that variety and heterogeneity adds ``x-factor'' to a speech and determines how engaged and appealing the speech is to an audience. Based on that, we chose 3 positive and 3 negative rating labels which cannot be otherwise trivially attributed to common identifiable causes. Note that the use of the term ``quality" associated with $HEM$ is in essence to emphasize the ``impact" of a speech on the audience. We do not claim $HEM$ (which represents ``heterogeneity" in a speech) is the sole or best way to capture ``effective influence" of a speech on audience but it definitely is one useful and relevant way of doing so (Figure~\ref{fig:HEM_all}).

\subsection{Relationship between $HEM$ and Ratings}
We first investigate the relationship between $HEM$ and ratings. We normalized the $HEM$ values using min-max technique such that they lie in $[0,1]$. These values are then divided into $5$ equal sized bins and their corresponding mean rating is reported in the X-axis of Figure \ref{fig:HEM_all}. The Y-axis denotes the corresponding rating value.\\ 
$HEM$ shows concave or increasing trend for positive rating labels (here, \emph{ingenious, fascinating, jaw-dropping}) and convex or decreasing trend for negative rating labels (here, \emph{long-winded, okay, unconvincing}). A greater value of $HEM_{tr}$ would indicate use of multiple characteristics of a good speech. This should potentially increase the \emph{fascinating} rating and decrease the \emph{unconvincing} rating. However, too much variety can be detrimental and make the talk annoying to viewers thereby decreasing \emph{fascinating} rating and increasing \emph{unconvincing} rating (Figure~\ref{fig:HEM_all}). This establishes that the psychological intuition motivated novel $HEM$ metric captures meaningfully influence of `` speech heterogeneity" on ratings for both the verbal and non verbal regime. The probability mass of distribution of $HEM_{tr}$ and $HEM_{ges}$ is small for higher and lower values respectively (Figure \ref{fig:HEM_dis} in Appendix ) causing comparably bigger errorbars around those regions. It has been shown that \cite{bolukbasi2016man} use of word embeddings may introduce or even amplify bias. However, in our case the HEM metric solely depends on the content and topics of the speech, irrespective of gender or race of the speaker. Any bias in word embedding if exists, will impact different genders and races equally. Moreover, the distribution of HEM for male and other genders (and white vs. others) speakers as shown in Figure 8 match closely confirming that the content of the speeches are not biased w.r.t gender of the speaker.   Also note that the normalized rating values are small because of the nature of distribution of ratings across all talks  (Figure \ref{fig:rating-dist} in Appendix).

\subsection{Bias in Rating Captured by $HEM$}\label{hem-bias}
\begin{figure*}[t]
    \centering
    \includegraphics[width=\textwidth]{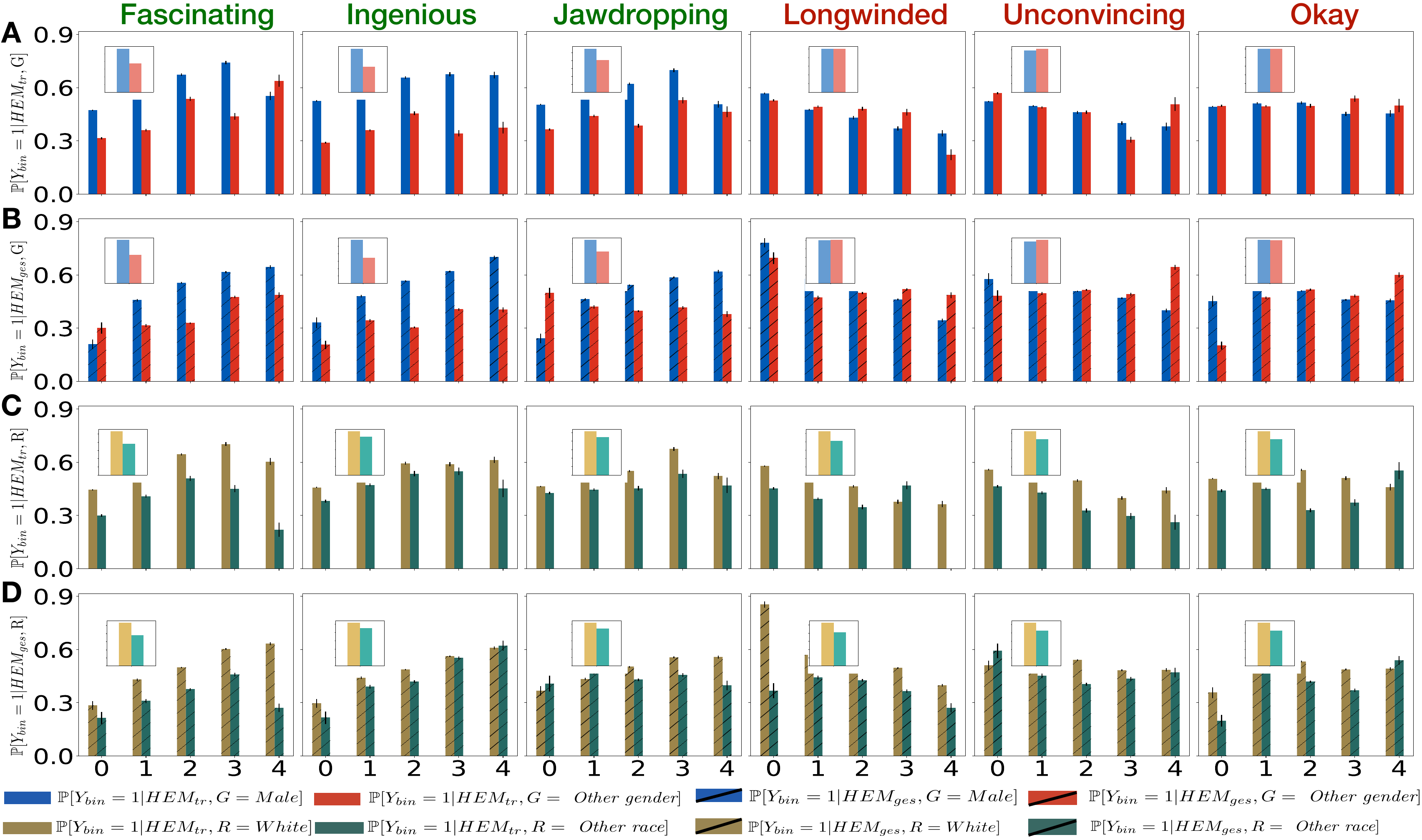}
    \caption{\textbf{Significant bias in rating w.r.t gender and $HEM$}\\ 
    \textbf{A:} Significant difference between probability of obtaining a rating for male speakers than for speakers of other gender w.r.t $HEM^{dis}_{tr}$ (trend agrees with true data in Figure \ref{fig:data-bias} A). Negative rating labels do not show significant bias in ratings as observed in true data. \textbf{B:} Same as \textbf{A} but for $HEM^{dis}_{ges}$. $HEM$ \textbf{C:} Significant difference between probability of obtaining a particular rating for white speakers than for speakers of other race w.r.t $HEM^{dis}_{tr}$ (trend agrees with true data in Figure \ref{fig:data-bias} B). \textbf{D:} Same as \textbf{C} but for $HEM^{dis}_{ges}$. The non overlapping 95\% CI shown in these plots indicate significance of our claim ($p<0.05$). }
    \label{fig:HEM_together}
\end{figure*}
The ratings of TED talks are given by spontaneous visitors to the website who come from different background and are of different age groups. We show that there exists bias in ratings when measured w.r.t race and gender. We find discrepancy between $\mathbb{P}[Y_{bin}=1| \text{ G } = \text{ Male }]$ and $\mathbb{P}[Y_{bin}=1| \text{ G } = \text{Other gender }]$ for a rating of interest $R$ as show in Figure \ref{fig:data-bias} A. Similar results for race are shown in  Figure \ref{fig:data-bias} B. It is important to solely draw attention to the difference in rating probabilities for sensitive attributes in Figure \ref{fig:data-bias} and not individual probability magnitudes. Bias in data can often be counter intuitive as shown in Figure \ref{fig:data-bias} B. It might be expected that white speakers get less negative ratings compared to speakers of other race. However, the data shows opposite trend, highlighting that bias is any type of discrepancy in rating probability irrespective of assumptions about superior race or gender. We investigate two cases in Figure \ref{fig:data-bias} to establish the existence of bias in ratings. Other combinations of sensitive attributes only make our claim about prevalent bias in data stronger.\\
We next tested whether $HEM$ captures such discrepancy of TED talk ratings w.r.t gender and race. Since $HEM$ captures the ``heterogeneity based quality" of the talk, the ratings should be similar for similar quality values, irrespective of race and gender. $HEM \in [0, 1]$ is dicretized into 5 values, $HEM^{dis} \in {0, 1, 2, 3, 4}$ by binning $[0, 1]$ into into 5 equal sized bins. We then compute $\mathbb{P}[Y_{bin}=1| \text{ G } = \text{ Male },HEM^{dis}]$ and $\mathbb{P}[Y_{bin}=1| \text{ G } = \text{ Other gender },HEM^{dis}]$ for a fixed value of  $HEM^{dis}$ as shown in Figure \ref{fig:HEM_together}. We find that male and female speakers have significant discrepancy across all positive rating labels for both transcript and gesture shown in Figure \ref{fig:HEM_together}. Note that the nature of the bias w.r.t. $HEM^{dis}$ matches existing bias in data. For example, true data shows that male speakers get ingenious rating with greater probability than speakers of other genders (for both transcript and gesture). This trend is nicely captured when computed w.r.t. $HEM$ values. Interestingly, we also observe less discrepancy for negative rating labels which is consistent with observation in true data. The distribution of $HEM$ values is similar within gender and race ensuring that the metric itself is not biased (Figure \ref{fig:hem-distribution} in Appendix). It is also important to notice that the bias has a definite pattern across almost all values of $HEM^{dis}$ for a particular rating and sensitive attribute. For example, when comparing w.r.t gender for the rating label jaw dropping, we observe that a male speaker is more likely to obtain that rating across all five values of $HEM^{dis}$. However, we see that for fascinating rating label the trend of bias in rating flips for  $HEM^{dis} _{tr} = 4$ as compared to other values of $HEM^{dis} _{tr} = {0, 1, 2, 3}$. Similarly, for the rating label jaw-dropping, we find a flip in trend of bias for $HEM^{dis} _{ges} = 0$ when compared to trend $HEM^{dis} _{ges} = {1, 2, 3, 4}$. This mismatch in bias pattern for higher $HEM^{dis} _{tr}$ and lower $HEM^{dis} _{ges}$ does not nullify the novelty of $HEM$ but can be explained by negligible probability mass of $HEM^{dis} _{tr}$ and $HEM^{dis} _{ges}$ for values of 4 and 0 respectively (see Figure \ref{fig:HEM_dis} in Appendix).\\ 
We further design a neural network model whose loss function enforces the reduction of rating differences for similar $HEM$ values. Note that the neural network has information about race and gender in the input but the rating discrepancy is minimized only w.r.t. the HEM metric only and explicitly.

\section{Fair Rating Prediction Model based on multi-modal $HEM$}

\subsection{Problem Formulation and Methods} 
Each TEDtalk's verbal feature is a $200$ dimensional vector obtained from the doc2vec representation~\cite{mikolov2013efficient}. To get the non-verbal feature we compute top $k$ (in our case $k=2$) eigenvalues for the correlation matrix of a segment (See Section \ref{section-hem} for detail about segment) and concatenate these eigenvalues across all the segments. We make this into a fixed length vector by padding $0$'s at the end. Other than the verbal and non-verbal features we also have the speaker's race and gender information.  Input $X^{(i)} (i \in \{1,\cdots,N\})$ is now created by concatenating the verbal feature, non-verbal feature  and the sensitive attributes (race and gender) and view counts. We use the binarized ratings $Y^{(i)}_{bin} = \left(y^{(i)}_1,\cdots, y^{(i)}_{6}\right)$ as the label. Also for each TEDtalk we have calculated its $HEM$ metric as defined in Section \ref{section-hem}. Let us denote the overall heterogeneity by $h^{(i)} = (HEM^{(i)}_{tr},HEM^{(i)}_{ges})$.\\ 
\begin{figure*}[t!]
    \centering
    \includegraphics[width=\textwidth]{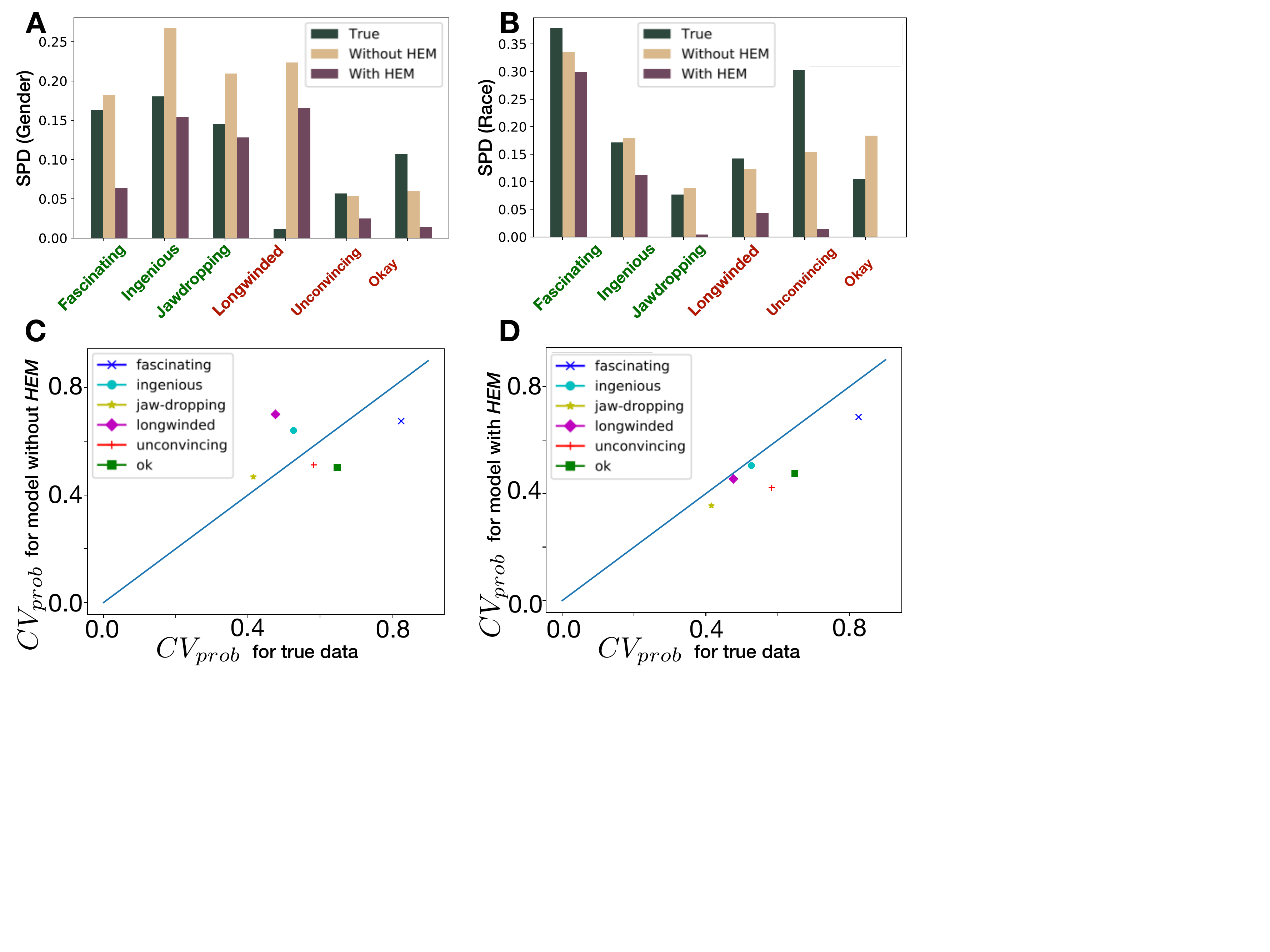}
      \caption{\textbf{Significant improvement of fairness by using modified loss function incorporating $HEM$}\\ \textbf{A:} Fairness in prediction improves as quantified by lower values of $SPD$ measure for gender. \textbf{B:} Same as \textbf{A} but for race. \textbf{C:} Fairness implies lower value of $CV_{prob}$ for the model as compared to true data. $CV_{prob}$ for three rating labels lie above the identity line indicating bias in prediction when $HEM$ is not incorporated in the loss function. \textbf{D:} Improvement in fairness w.r.t $CV_{prob}$ for all 6 rating labels for model with $HEM$ in the loss function. }
    \label{fig:fairness-spd}
\end{figure*}
Our goal is to learn a prediction  function $f_{\theta}$ such that its predicted output is not only close to the ground truth but also has similar prediction for inputs with similar $HEM$ metric value. Formally we need to minimize the following loss function,
\begin{equation}\label{opt-func}
    L(\theta) = Pred(\theta)+\lambda \cdot HEM(\theta)
\end{equation}
where, $Pred(\theta)=\frac{1}{N}\sum_{i=1}^N BCE\left(f_{\theta}(X^{(i)}) , Y^{(i)}\right)$ and $HEM(\theta)=\frac{1}{\binom{N}{2}}\sum_{i,j}\norm{\hat{Y}^{(i)}-\hat{Y}^{(j)}}^2_2 \cdot \mathbb{I}\left(|h^{(i)} -h^{(j)}|<\epsilon\right)$
\noindent where $BCE$ denotes the binary cross entropy loss and $\hat{Y}^{(i)} = f_{\theta}(X^{(i)})$. Here $Pred(\theta)$ represents the prediction loss w.r.t the ground truth and $HEM(\theta)$ controls the discrepancy of the prediction between inputs with similar HEM metric. Here $\epsilon$ and $\lambda$ are the hyperparameters, where $\epsilon$ is the tolerance of $HEM$ difference for which discrepancy in rating is penalized  and $\lambda$ controls the strength of $HEM$ loss. Higher $\lambda$ forces similar rating for talks with similar $HEM$. We use a feed-forward neural network with 1 hidden layer to learn function $f_{\theta}$ which minimizes the loss defined in \eqref{opt-func}. \\
\noindent \textbf{Remark:} The loss function does not have any explicit component for reducing unfairness w.r.t sensitive attributes (race and gender). Still minimizing this loss function improves fairness as we can see in the next section.

\subsection{Results} 
We have used a feed forward neural network with one hidden layer of 400 hidden nodes for classification. From \eqref{opt-func} we can see that in case of $\epsilon = \lambda = 0$ the model does not use the knowledge of $HEM$ metric, hence it only focuses on accuracy. With the increase of $\epsilon$ and $\lambda$ it aims to improve fairness which leads to reduction in model performance as a trade-off. One challenge is to find a good set of $\epsilon$ and $\lambda$ which optimizes both accuracy and fairness. We performed a grid search to find the suitable hyperparameters: $\epsilon$ in the range $0.01$ to $0.2$ and $\lambda$ between $0.2$ and $10$ (see Table \ref{modelaccuracytab} in Appendix). We use binary accuracy (percentage of data correctly classified) to measure the performance of the model and $SPD$~\cite{biddle2006adverse} to quantify the fairness, where $SPD$ measures the difference of rating predictions between two groups. Formally,
$\textrm{SPD} = |\mathbb{P}(y_i = 1 | S \in G_1)-\mathbb{P}(y_i=1|S \in G_2)|$. We measure $SPD$ both w.r.t. gender (i.e. $G_1 =$  Male and $G_2$ = other gender) and race (i.e. $G_1 =$  White and $G_2$ = other race). Smaller values of SPD indicates better fairness. Note that the SPD decreased after incorporating HEM into the loss function which indicates fair rating prediction. \\
The average accuracy of all 6 rating labels is 67\% when we train our model without any $HEM$ loss ($\epsilon = \lambda = 0$). After adding the $HEM$ loss to our model (with $\epsilon=0.017$ and $\lambda=5.0$), we achieve 64\% accuracy on rating prediction but the fairness w.r.t. gender improved significantly as shown in Figure \ref{fig:fairness-spd} A. For  the choice of $\epsilon=0.02$ and $\lambda=4.0$ we observe similar model performance (64\% accuracy) and fairness w.r.t. race (Figure \ref{fig:fairness-spd} B). Note that, the model's accuracy is decreased by 3\% only, when trained with $HEM$ in the loss function. The goal of this work is not to beat state of the art in prediction accuracy but to achieve improved fairness in prediction for comparable accuracy levels.\\ 
In general, the $SPD$ metric quantifies fairness in prediction between two complementary groups (e.g. \textit{male} and \textit{not male}). In our dataset, there are $3$ gender labels and $4$ race labels, allowing us to measure fairness of the model's prediction across 12 different pairs of groups. Previous work on fairness in public speech introduced a novel metric, $CV_{prob}$, to collectively quantify fairness across all these 12 groups by measuring their variability \cite{acharyya2020fairyted}. Intuitively, $CV_{prob}$ compares variability of ratings across possible instances of sensitive attributes for the prediction model with and without incorporation of fairness. For example, let us assume that White and African American speakers have probabilities of 0.6 and 0.3 to be rated \textit{fascinating}. It would be expected that after incorporation of fairness into the prediction model this variability in rating probabilities will drop, becoming 0.58 and 0.55 say. Hence lower the variability of rating probabilities across 12 groups as compared to true data, lower will be the values of $CV_{prob}$ for the fair model and higher will be the fairness in prediction (details about the metric can be found in Section 7.3 of \cite{acharyya2020fairyted}). We show (Figure \ref{fig:fairness-spd} C and D) that our model when trained with $HEM$ loss reduces the variability in rating probability across 12 groups for each of the 6 ratings considered (all dots lie below identity line) as compared to true data. It is important to note that $CV_{prob}$ for model with $HEM$ in loss function not only lies below identity line but has lower magnitude for all 6 ratings as compared to $CV_{prob}$ for model without $HEM$ in loss function (compare y-axis values for C and D in Figure \ref{fig:fairness-spd}), indicating increase in fairness across all rating labels. Code for reproducing results is available.\footnote{\url{https://bit.ly/2MFhUdB}}

\section{Conclusion}
In this work we take the following steps to build a fair rating prediction model for public speeches: 1) define a heterogeneity metric $HEM$, that quantifies the quality of a talk based on variation of characteristics in transcript and gestures used by a speaker,2) show $HEM$ has meaningful relationship with viewer ratings for TED talks 3) show that $HEM$ captures the bias in ratings for public speeches  4) define a loss function by incorporating $HEM$ to design a fair rating prediction model. Though our focus was mostly on verbal and non-verbal aspects of the speeches, it can be extended to other modes due to simplicity of our encoding systems for defining $HEM$.\\ 
In today's world, where new talent and careers boom on social media and the web, unfair rating systems can harm career growth or even add to imposter syndrome in beginners. With careful and exhaustive investigation and exploration, our work can be extended and strengthened in various domains with the goal of preventing adverse effects of unfair ratings available on online platforms. Often in hiring systems, ratings from multiple interviewers are used to make the final decision and our model can be a starting point for such applications in future work. 
The simplicity of our model would also allow extensions and future work on determining fair ratings for job talks, podcasts etc.
Our work exploits multi-modality of dataset and draws inspiration from psychological studies to incorporate fairness in rating prediction systems. Future studies can be inspired from this, to explore a wide range of factors that introduce bias in our prediction models and thereby encourage building of applications that positively impact the society. 

\bibliography{anthology,custom}
\bibliographystyle{plain}

\appendix
\onecolumn
\clearpage
\section{Tables}

\begin{table}[h]
  \caption{Breakdown of TED talk Dataset w.r.t. gender and race}
  \begin{center}
  \small
\scalebox{0.8}{
  \begin{tabular}{|c|c|c|}
    \hline
    \textbf{Property}& \textbf{Sub Property} &\textbf{Quantity}\\
    \hline
    Total Number of talks & &1980\\
    \hline
    & Male& 1307 \\
    Gender of Speaker & Female & 659\\
    & Other gender & 14\\
    \hline
     & White & 1573\\
     Race of Speaker & African American & 147\\
     & Asian & 173\\
     & Other Race & 87\\
    \hline
\end{tabular}}
\end{center}
\label{tab:datasize}
\end{table}
\begin{table}[h]
  \caption{Pre-processed TED Dataset Attributes}
  \begin{center}
  \small
\scalebox{0.8}{
  \begin{tabular}{|c|c|}
    \hline
    \textbf{Sensitive attributes $S$}& $R$: race and $G$: gender\\
    \hline
    \textbf{Data attributes $X$} & $Tr$: transcript $V$: visual gestures \\
    & and $C$: view count\\
    \hline
    \textbf{Label $Y$} & ratings (3 positive and 3 negative)\\
    & and $Y$: normalized ratings,\\
    &$Y^{bin}$: binarized ratings\\
    \hline
\end{tabular}}
\end{center}
\label{tab:final_data}
\end{table}

\begin{table}[h]
\caption{Results of TEDtalk rating prediction. Accuracy is the binary accuracy of the classification model. $\epsilon$ and $\lambda$ are the hyperparameters as defined in \eqref{opt-func}. M-SPD (Gender) denotes SPD of the predicted label w.r.t. gender.  M-SPD (Race) similarly defined for race. True-SPD (Gender) denotes the SPD of the true dataset w.r.t. gender. Lower values of SPD implies better fairness. It is to see that SPD of the predicted labels is best when the model is trained using HEM loss. Best (lowest) SPD w.r.t. race is achieved across all ratings when trained with HEM. In case of \textit{longwinded} rating the SPD w.r.t. gender has increased from true SPD. However, this is much lower than the  M-SPD when trained without HEM loss.}

\label{modelaccuracytab}
\small
\scalebox{1}{
\begin{tabular}{|l|c|c|c|c|c|c|}  
\hline    
Ratings &  $(\epsilon,\lambda)$ &  Accuracy $\uparrow$ & M-SPD (Gender) $\downarrow$ & True SPD (Gender) &  M-SPD (Race) $\downarrow$ & True SPD (Race)  \\  
\hline 
fascinating &      (0,0) &          0.77 &  0.18  &  0.16 &        0.33 & 0.38\\
 &      (0.02,4) &          0.77 &  0.16  &   &        0.30 & \\
 &      (0.017,5) &          0.75 &  $\bm{0.06}$  &   &        $\bm{0.19}$ & \\
 \hline
ingenious &      (0,0) &          0.76 &  0.27  &  0.18 &          0.18 & 0.17 \\ 
 &      (0.02,4) &          0.72 &  0.15  &   &          $\bm{0.11}$ &   \\ 
 &      (0.017,5) &          0.73 &  $\bm{0.15}$  &   &          0.29 &   \\ 
 \hline
jaw-dropping &      (0,0) &          0.69 &  0.21  &  0.15 &          0.09  & 0.08 \\
 &      (0.02,4) &          0.65 &  $\bm{0.09}$  &   &          $\bm{0.004}$  &  \\
 &      (0.017,5) &          0.61 &  0.13  &   &          0.18  &  \\
 \hline
longwinded &      (0,0) &          0.61 &  0.22   & 0.01 &          0.12 &  0.14 \\
 &      (0.02,4) &          0.55 &  $\bm{0.12}$   &  &          $\bm{0.04}$ &   \\
 &      (0.017,5) &          0.59 &  0.17   &  &          0.17 &   \\
 \hline
unconvincing &      (0,0) &          0.61 &  0.05   & 0.06 &          0.15  & 0.30 \\ 
&      (0.02,4) &          0.59 &  0.07   &  &          $\bm{0.01}$  &  \\ 
 &      (0.017,5) &          0.57 &  $\bm{0.02}$   &  &          0.02  &  \\
 \hline
ok &      (0,0) &          0.61 &  0.06   &  0.11 &          0.18  & 0.10 \\
 &      (0.02,4) &          0.58 &  0.01   &   &          $\bm{0.001}$  &  \\
 &      (0.017,5) &          0.61 &  $\bm{0.01}$   &   &          0.13  &  \\
\hline
\end{tabular}
}
\end{table}

\newpage
\section{Extra Figures}

\begin{figure}[h]
    \centering
    \includegraphics[width = \columnwidth]{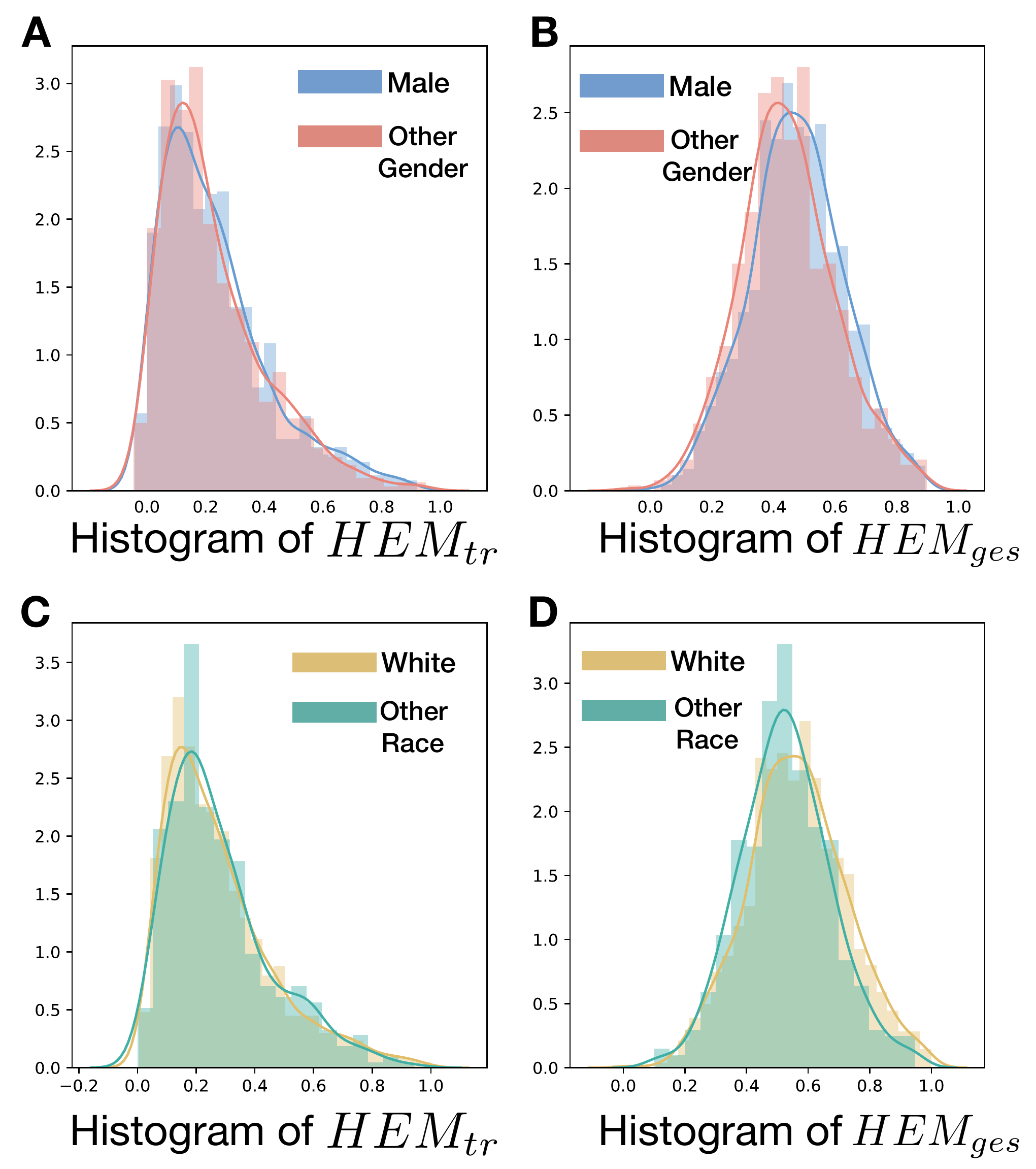}
    \caption{$HEM_{tr}$ and $HEM_{ges}$ are not biased by definition\\
    A-B: Shows histogram of $HEM_{tr}$ and $HEM_{ges}$ for male speakers (blue) and speakers of other gender (red). The almost overlap between the two histograms indicate that the bias observed in Figure \ref{fig:HEM_together} are genuine and not an artifact due to a biased metric. C-D: Same as A-B but for race comparing white speakers (yellow) with speakers of other gender (green) for both $HEM_{tr}$ in C and $HEM_{ges}$ in D. We find no significant difference in the two histograms.}
    \label{fig:hem-distribution}
\end{figure}

\begin{figure}
    \centering
    \includegraphics[width = \columnwidth]{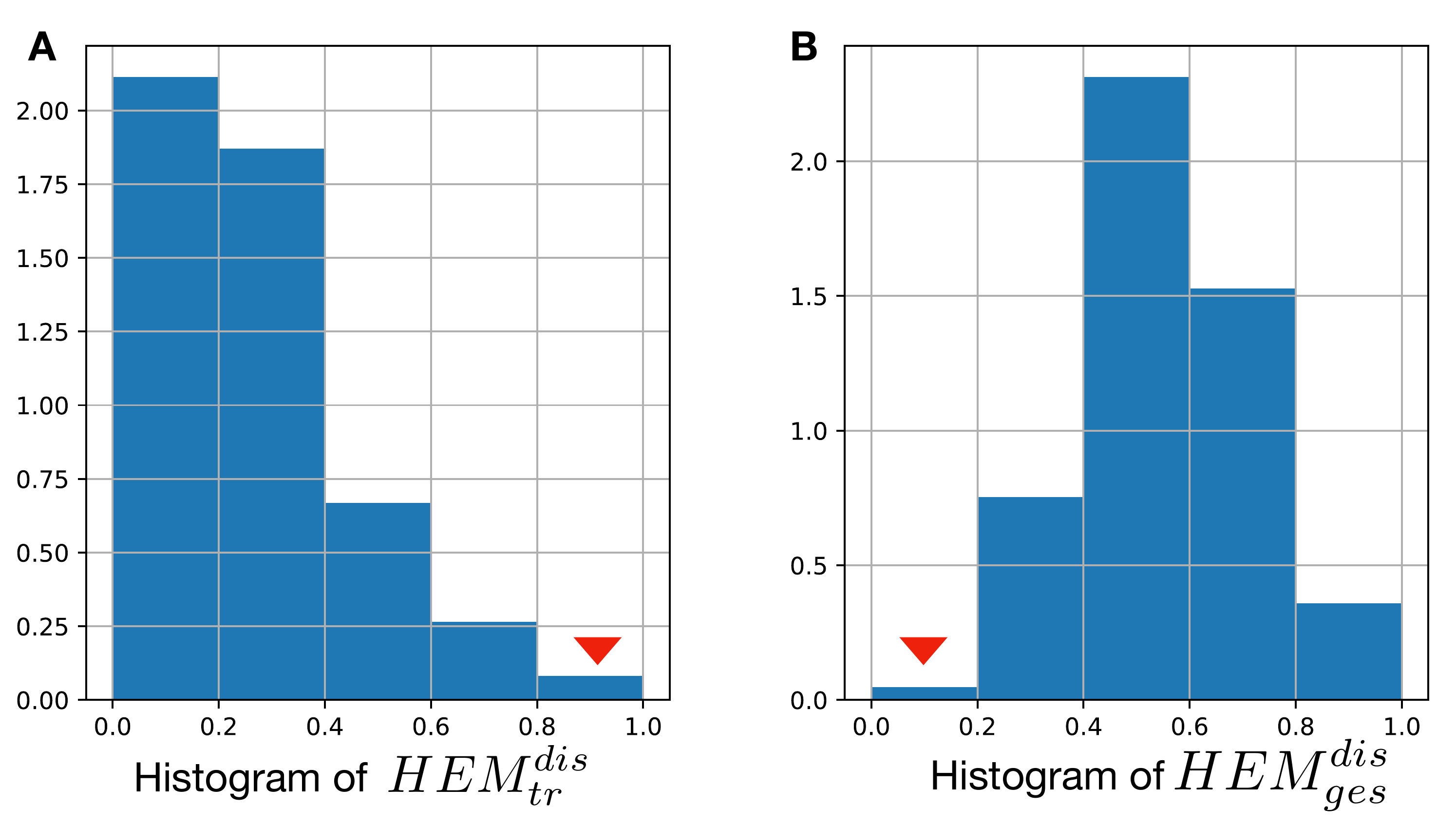}
    \caption{Distribution of discretized $HEM^{dis}_{tr}$ and $HEM^{dis}_{ges}$\\
    A: $HEM_{tr}$ is discretized by assigning values between $[0.0, 0.2)$ to category $0$, $[0.2, 0.4)$ to category $1$ and so on $[0.8, 1.0]$ to $4$. We find negligible instances for $HEM^{dis}_{tr} = 4$, (i.e, $HEM_{tr}>=0.8 \text{ and} <=1.0$) as shown in red.  B: Similarly, $HEM^{dis}_{ges} = 0$, (i.e, $HEM_{ges}>=0.0 \text{ and} <0.2$) is almost non-existent, pointed by red arrow. }
    \label{fig:HEM_dis}
\end{figure}

\begin{figure}
    \centering
    \includegraphics[width=\columnwidth]{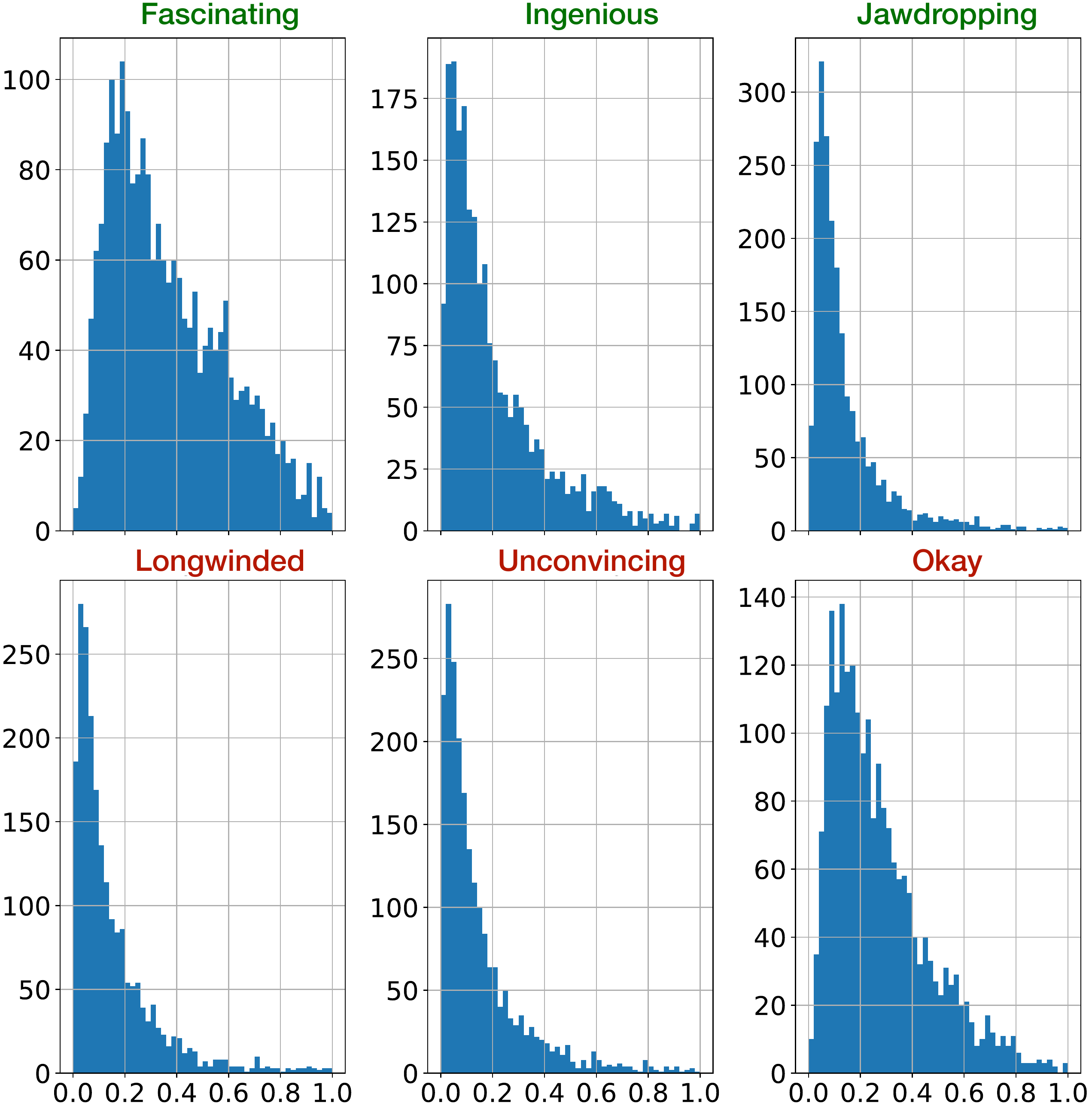}
    \caption{Distribution of non-binarized rating values. Higher mass on smaller values of ratings justifies the small change of rating w.r.t. HEM metric in Figure \ref{fig:HEM_all}.}
    \label{fig:rating-dist}
\end{figure}

\begin{figure}
    \centering
    \includegraphics[width = \columnwidth]{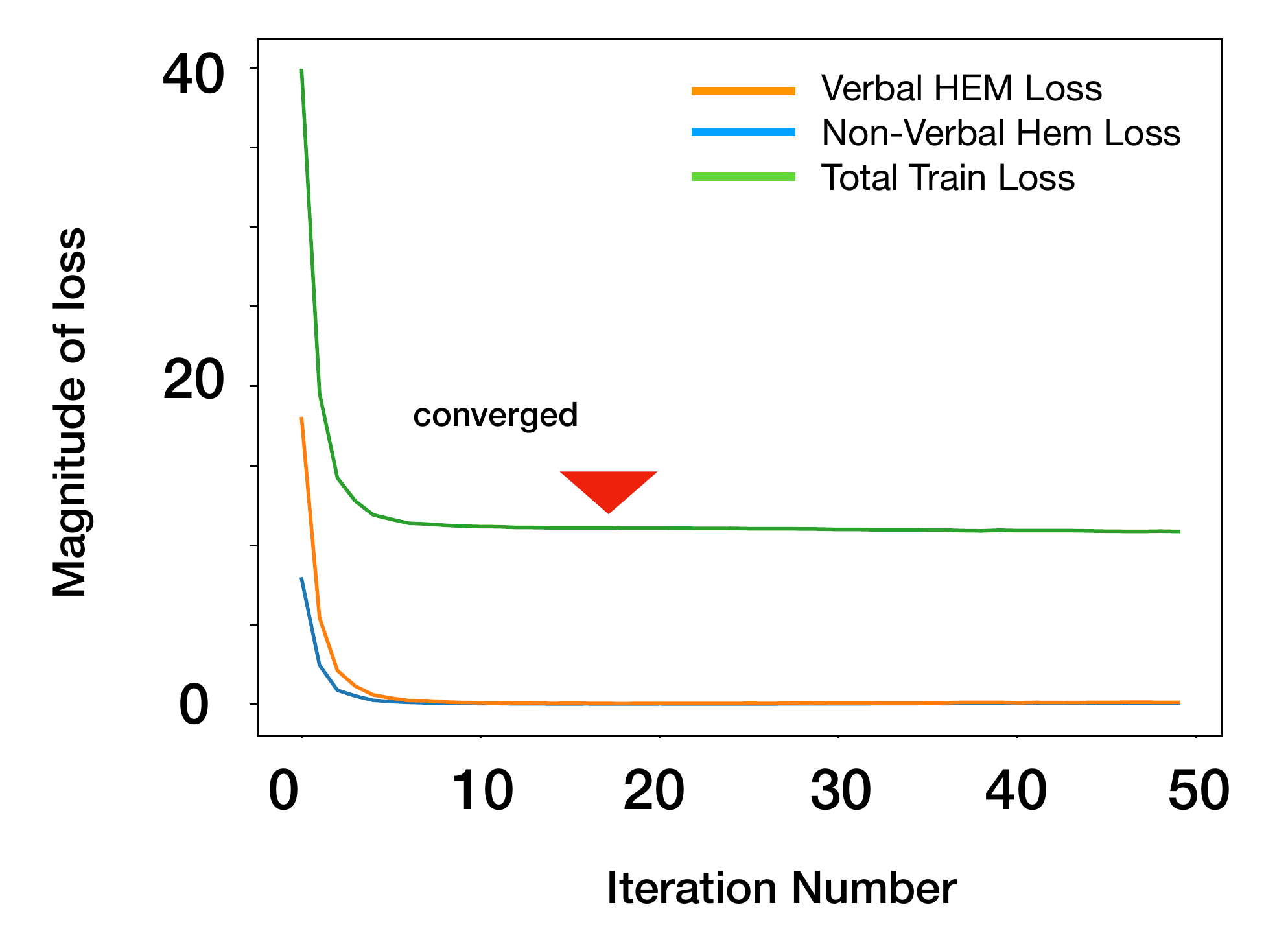}
      \caption{Convergence in training neural network\\
      The loss function decreases and finally converges with increase in iteration steps indicating that learning is complete in the neural network. We show training loss w.r.t $HEM_{tr}$ in orange, $HEM_{ges}$ in blue and total training loss in green.}
    \label{fig:loss}
\end{figure}

\end{document}